\documentclass[conference]{IEEEtran}
\usepackage[utf8]{inputenc}
\usepackage{dtk-logos} 

\usepackage{color}
\usepackage{amsmath}
\usepackage{graphicx}
\usepackage{multirow}
\usepackage{booktabs}
\usepackage [table]{xcolor}
\usepackage{color}
\usepackage{amsmath}
\usepackage{graphicx}
\usepackage{multirow}
\usepackage{booktabs}
\usepackage{epstopdf}
\usepackage [table]{xcolor}
\usepackage{amsmath}
\usepackage{graphicx}
\usepackage{booktabs}
\usepackage[table]{xcolor}
\usepackage{pdflscape}
\usepackage{amsmath}
\usepackage{graphicx}
\usepackage{booktabs}
\usepackage{balance}

\usepackage{color}
\usepackage{subcaption}
\usepackage{graphicx}
\usepackage{multirow}
\usepackage{booktabs}
\usepackage[table]{xcolor}
\usepackage{xcolor, colortbl}
\usepackage{hhline}
\usepackage{bbding}
\usepackage{enumerate}
\usepackage{amsmath}
\usepackage{color}
\usepackage{amsmath}
\usepackage{graphicx}
\usepackage{float}
\usepackage{multirow}
\usepackage{booktabs}
\usepackage{epstopdf}
\usepackage [table]{xcolor}
\usepackage{algorithmic}
\usepackage{pseudocode}
\usepackage[]{algorithm2e}
\usepackage{amsmath}
\usepackage{caption}
\usepackage{subcaption}
\usepackage[utf8]{inputenc}
\usepackage[english]{babel}
\usepackage{float}
\usepackage{wrapfig}
\usepackage{multicol, blindtext}
\usepackage{lipsum}

\title{IEEE \BibTeX{} style}
\author{ShareLaTeX Templates}
\date{January 2014}

\begin{document}

\title{Evaluating the Communication Efficiency in Federated Learning Algorithms}

\author{\IEEEauthorblockN{ Muhammad Asad, Ahmed Moustafa, Takayuki Ito, Muhammad Aslam$^{\dag}$}
\IEEEauthorblockA{Department of Computer Science, Nagoya Institute of Technology, Nagoya - Japan \\
$^{^\dag}$School of Cyber Science and Engineering, Wuhan University, Wuhan - China \\
e:mail: m.asad@itolab.nitech.ac.jp
}}

%

\maketitle

\begin{abstract}

In the era of advanced technologies, mobile devices are equipped with computing and sensing capabilities that gather excessive amounts of data. These amounts of data are suitable for training different learning models. Cooperated with advancements in Deep Learning (DL), these learning models empower numerous useful applications, e.g., image processing, speech recognition, healthcare, vehicular network and many more. Traditionally, Machine Learning (ML) approaches require data to be centralised in cloud-based  data-centres. However, this data is often large in quantity and privacy-sensitive which prevents logging into these data-centres for training the learning models. In turn, this results in critical issues of high latency and communication inefficiency. Recently, in light of new privacy legislations in many countries, the concept of Federated Learning (FL) has been introduced. In FL, mobile users are empowered to learn a global model by aggregating their local models, without sharing the privacy-sensitive data. Usually, these mobile users have slow network connections to the data-centre where the global model is maintained. Moreover, in a complex and large scale network, heterogeneous devices that have various energy constraints are involved. This raises the challenge of communication cost when implementing FL at large scale. To this end, in this research, we begin with the fundamentals of FL, and then, we highlight the recent FL algorithms and evaluate their communication efficiency with detailed comparisons. Furthermore, we propose a set of solutions to alleviate the existing FL problems both from communication perspective and privacy perspective. 

\end{abstract}


\section{Introduction}

In the past few years, the number of intelligent devices has grown rapidly with the advent of Internet of Things (IoT) \cite{wiedemann2019compact}. These devices are able to collect and process data at exceptional scale because of their embedded sensors and potent hardwares. On the other hand, Deep Learning (DL) has transformed the ways of information extraction from the data sources with radical successes in many applications such as image processing, speech recognition, health care, and natural language processing (NLP). The astonishing success of DL in processing large amounts of data can be credited to the availability of sufficient datasets for training \cite{mcmahan2016communication}. In this regard, IoT has enabled a significant improvement in the training of DL models by exploiting the huge amounts of recorded data. Meanwhile, privacy has emerged as a major concern for each mobile user and it grows rapidly with the advent of social media networks. In this context, multiple misuse and data leakage cases in recent times has demonstrated that users' privacy is at high risk during the centralised processing of data \cite{mcmahan2016communication}. 

Usually, IoT devices collect data in private environments where each device is explicitly decoupled from users due to various reasons. Therefore, sharing this data with a centralised server is not a good option, and hence, the possibility of training a DL model becomes challenging. To address this dilemma, a decentralised Machine Learning (ML) approach has been introduced, namely Federated Learning \cite{mcmahan2016federated}.  

Federated Learning (FL) allows each participant device to jointly train a global DL model by using their combined data without revealing the personal data of each device to the centralised server. This privacy-preserving collaborative learning technique is achieved by following a three-step process as illustrated in Figure \ref{FL}.

\begin{itemize}

\item \textbf{Task Initialization:} From the thousands of available devices in a certain time, the centralised server selects a certain number of devices and decides the training of specific FL task i.e., according to the corresponding data and the target application. Then, the centralised server specifies the training process and the hyper-parameters, e.g., learning rate. After specifying the devices and other task requirements, the centralised server broadcast the global model and the FL task to the selected participants. 

\begin{figure}
\begin{center}
 \includegraphics [width=3.2in,height=1.8in]{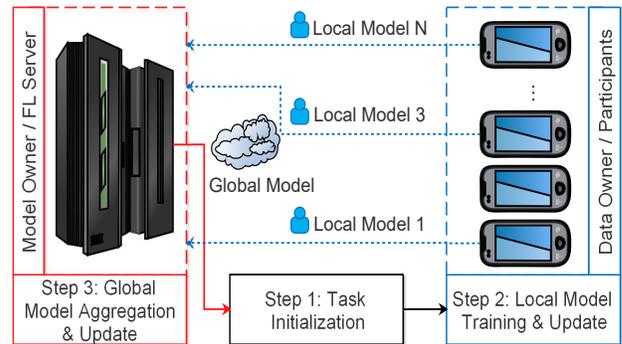}
    \centering
\vspace{-7pt}
  \caption{FL general process with server and participants in single communication round where participants synchronise with the server and update the server for new updated global model.} \label{FL}
 \end{center}
\vspace{-17pt}
\end{figure}

\item \textbf{Local Model Training:} After receiving the global model, all the participants perform local computation based on the global model to update thier local parameters. The purpose is to find the optimal parameters that help in reducing the loss function. The updated parameters of each local model is sent back to the centralised server. 

\item \textbf{Global Model Aggregation:} The centralised server receives the local parameters from each participant and updates the global model parameters and then send back the updated global model parameters to all the participants in order to reduce the global loss function.  
\end{itemize}

The process of training in Steps 2 and 3 is continuously repeated until the desirable training accuracy is achieved or the global loss function meets the minimum requirements \cite{lim2019federated}.

On the other hand, in conventional centralised ML models, the implementation of federated training approaches on mobile networks features the following advantages. Firstly, \textit{efficient use of network bandwidth:} data owners only send the updated model parameters instead of sending the raw data for aggregation, which reduces the significant cost of data communication \cite{reisizadeh2019fedpaq}. Secondly, \textit{latency:} in time critical applications such as Intelligent Transport Systems (ITS) where minimum network delay can create threatening situations, the implementation of FL can minimise this delay as the ML models will be consistently trained and updated. Moreover, real-time decisions, e.g., event detection, can be made locally at end devices \cite{ang2019robust}. Consequently, the latency in FL systems are much less than centralised systems. Thirdly, \textit{privacy:} the participants are not sending their raw data to the centralised server which ultimately guarantees each user privacy, and with this guaranteed privacy, the maximum number of users is able to participate in collaborative model training, and hence, the built model becomes better \cite{nasr2019comprehensive}. 

Up to date, many attempts have been made for the implementation of FL at scale but still there are several challenges that need to be considered. Firstly, from the perspective of resource allocation, the heterogeneity of participating devices in terms of computation power, data quality and participation rate, needs to be managed in large scale networks. Secondly, due to the limited communication bandwidth and high dimensional model updates in mobile devices, communication cost remains an issue. Thirdly, malicious participants may exist in FL and can share the parameters of other participants, therefore, the security and privacy issues need to be considered in depth. 

In the existing approaches of FL, neither the communication issues are properly addressed, nor the challenges of FL implementation are deeply discussed. This motivates us for conducting this study that covers, 1) implementation of FL 2) communication cost, and 3) a statistical and experimental comparison of the existing state-of-the-art algorithms. For the reader's convenience, a list of the common abbreviations is shown in Table I.


The rest of this paper is organised as follows. Section II, highlights the fundamentals of FL and provides an overview of the existing FL frameworks and algorithms. Communication efficiency in FL algorithms is discussed in Section III. Evaluation of these algorithms are given in Section IV. The future work is proposed in Section V. Section VI concludes this paper.  

\begin{table}
\centering
\label{schetab}
{\small
\begin{tabular}{|p{1.8cm}|p{5.2cm}|}
\hline
   {Abbreviation} & {Description}      \\\hline
 \hline
CNN & Convolution Neural Network \\ \hline
DNN & Deep Neural Network \\ \hline
DL & Deep Learning   \\ \hline
ML & Machine Learning  \\ \hline
DRL & Deep Reinforcement Learning   \\ \hline
FL & Federated Learning  \\ \hline
FedAvg & Federated Averaging   \\ \hline
IID & Independent and Identically Distributed  \\ \hline
NLP & Natural Language Processing   \\ \hline
MLP & Multilayer Perceptron \\ \hline
ITS & Intelligent Transport Systems  \\ \hline
IoT & Internet of Things  \\ \hline
LSTM & Long Short Term Memory  \\ \hline
TFF & TensorFlow Federated  \\ \hline
SGD & Stochastic Gradient Descent  \\ \hline

  \end{tabular}
}
\label{Abbreviations}
\caption{List of Notations and Common Abbreviations}
\end{table}


\section{Fundamentals of Federated Learning}

In this section, we highlight the characteristics, challenges, existing frameworks and the state-of-the art algorithms of FL.  

\subsection{Federated Learning Challenges}

In this subsection, we identify the unique and challenging characteristics of FL which distinguish it from the conventional distributed learning approaches. These characteristics are listed below:

\begin{itemize}
\item \textbf{Non-IID Data:} data on the participant devices are collected by the devices themselves so there is a huge possibility of different data distributions among all the participants as each individual participant device collects the data based on its personal usage pattern and its local environment which might be different from other participants \cite{zhao2018federated}. 

\item \textbf{Number of Clients:} collaborative learning models are evaluated based on the number of participants and the available data which are essential for FL. In this context, client participation is a big challenge as clients often denies to participate in the training due to various reasons, e.g., poor connection, limited battery, or no interest in collaborative training \cite{nishio2019client}. Therefore, guaranteeing the participation in FL needs to be solved. 

\item \textbf{Parameter Server:} after exceeding a certain threshold, the increasing number of clients becomes infeasible because of linear growth in the workload of communication and aggregation. Therefore, it is much needed to communicate via a parameter server in FL. Using this parameter server, communication rounds reduce to single round for participants and the server. Moreover, it reduces the communication cost per client. However, communication through parameter server remains a challenge for communication-efficient distributed training because the upload and download to/from the server require efficient compression in order to reduce communication-cost, time and energy consumption.  

\item \textbf{Limitations in Battery and Memory:} clients in the FL are usually the mobile devices which often have limited battery capacities. However, each single iteration of Stochastic Gradient Descent (SGD) in order to train Deep Neural Networks (DNN) is quite expensive in terms of battery cost. Therefore, it is necessary to use a small number of iterations during SGD evaluation. Moreover, the size of memory on mobile devices are so limited that it might be impossible to memorise all the processed samples during training because the footprint of SGD grows linearly with batch size \cite{sattler2019robust}. 

\end{itemize}

In summary, the aforementioned characteristics of FL require consideration when designing communication-efficient distributed training algorithms. 

\subsection{Federated Learning Frameworks}

Among the various FL frameworks, some of the open-source frameworks are developed for the implementation of FL algorithms as follows: 

\begin{enumerate}

\item \textbf{TensorFlow Federated (TFF):} TFF is an open-source framework for ML and similar computation on decentralised data. TFF is developed by Google to enable researchers to experiment in FL. The interface of TFF consists of the following two layers; Federated Learning (FL) and Federated Core (FC).  

1) FL (API): in this layer, users are not required to apply their own FL algorithms instead it offers a high-level interface that allows users to implement FL using the existing models of TF. 

2) FC (API): in this layer, users can implement their personal FL algorithms using a lower-level interface which combines TF with distributed communication operators.  

In addition, developers are enabled to express federated-computation in TFF for diverse runtime-environments \cite{tensorfed}. 

\item \textbf{PySyft:} PySyft is an open source library based on PyTorch for the implementation of FL algorithms which allows developers to train their models in untrusted environments with complete features of security. Retaining the native Torch interface, PySyft execute all the tensor operations in a similar fashion as in PyTorch. Apart from the FL, PySyft leverages other useful techniques in ML e.g., secure multi-party computation and differential privacy \cite{ryffel2018generic}.  

\RestyleAlgo{boxruled}
\begin{algorithm}
    \SetKwInOut{Input}{Input}
    \SetKwInOut{Output}{Output}

\Input{Mini-batch size (B), Participants (k), Participants per epoch (m), Total epochs (E) and Learning rate $\eta$}
\Output{Global model $W_{GM}$}
\textbf{Server Execution:} \\
\textbf{Initialize} $W_{GM}$: \\

    \For {each epoch = $1,2,3 ... N$ }{
   Random subset $S_{t}$ of $m$ participants from $k$ participants \\
    }
      \For {every particpant $w$ $\in$ $S_{t}$ \textbf{parallely}}{
      ${w_{GM}}^{t+1}_k$ $\leftarrow$ \textbf{ClientTrainingUpdate}($k, w_{GM}$) 
    }
$w_{t+1}$  $\leftarrow$  $\sum_{1}^{k}$ $\frac{m_{k}}{m}$ ${w_{GM}}^{t+1}_k$ (Averaging Aggregation) \\
\textbf{Client Update:} \\
$\beta$ $\leftarrow$ mini-batches creates through splitting local datasets $D_L$ \\
\For {each local epoch $k$ from $1$ to $E$ }{
    }
\For {local mini-batch \textit{b} $\in$ $\beta$}{
$w_{GM}$ $\leftarrow$ \textit{w} -- $\eta$$\bigtriangleup$\textit{l}\textit{(w, b)} \\\
\textbf{\textit{($\bigtriangleup$l is the gradient of l on b and $\eta$ is the learning rate)}}
    }

    \caption{Federated Averaging Algorithm \cite{mcmahan2016communication}}
\end{algorithm}

\item \textbf{LEAF:} LEAF is an open-source benchmark framework for FL that includes numerous datasets, e.g., FEMNIST, Sentiment140, Shakespeare, Celeba and Synthetic. In Federated Extended MNIST (FEMNIST) and Sentiment140 datasets, partitions are based on writer-of-each-character and different-users, respectively. The participants on these datasets in FL are assumed to be a writer or a user and the corresponding data will remain on each participant's devices \cite{caldas2018leaf}. 

\end{enumerate}

\subsection{Federated Learning Approaches}

In a typical ML system, optimisation algorithms like SGD require large datasets for efficient training over the cloud. Such iterative algorithms demand high-throughput and low latency connection for training. In case of FL, data is distributed over millions of devices in a heterogeneous manner. Moreover, those devices have significantly lower-throughput and higher-latency connections and intermittently ready for training. Motivated by latency and bandwidth limitations, Federated Averaging Algorithm (FedAvg) is proposed \cite{mcmahan2016communication} to overcome these issues. Pseudocode of FedAvg algorithm is shown in Algorithm 1. 

FedAvg algorithm works as follows, firstly, the server initialises the task (server execution: \textbf{Algorithm 1}) and then the implementation of local training begins by participants using the mini-batches of local datasets. Secondly, participants optimises the task in client update (client update: \textbf{Algorithm 1}). Lastly, in the final iteration of \textbf{client update}, the global loss is minimised by the server using averaging aggregation which is generally defined as: 

\begin{equation}
w_{t+1}  =  \sum_{1}^{k} \frac{m_{k}}{m} {w_{GM}}^{t+1}_k 
\end{equation}

As described in Section 1, the training process of FL will continuously repeat until the desired accuracy is achieved or the global loss converges. 
Although FedAvg is specifically proposed for FL settings, it suffers with large amounts of non-iid data as observed in \cite{zhao2018federated} such that its accuracy drops by 55\% in non-iid environments as compared to iid environments. This can be more severe in highly distributed environments because of the different data distribution of each client \cite{li2019convergence}. Based on these analysis, we believe that the FedAvg algorithm faces a convergence issue on non-iid data. 

To reach the SGD level convergence rate, \textbf{\textit{SignSGD}} is proposed in \cite{bernstein2018signsgd} which transmits only the sign of each mini-batch in CNN and MLP. Compared with FedAvg, this method suffers even worse stability in non-iid environments \cite{bernstein2018signsgd1}. As demonstrated in \cite{sattler2019robust}, SignSGD completely fails in the accuracy test of CIFAR-10 dataset and the convex logistic regression test. To understand this convergence issue, we need to investigate how a single mini-batch can have a correct sign. We consider 

\begin{equation}
BG_{p}^{s} = \frac{1}{s} \sum_{i=1}^{s} \bigtriangledown_{p}l (x_{i}, \textit{Z}) 
\end{equation}

a batch gradient $BG$ on data $d^{s}$ from a specific mini-batch $mb$ $\subset$ $d$ of size $s$ with parameter $p$. If we consider this gradient $BG_{p}^{s}$ over the complete training data $d$. Then, the probability of the gradient can be defined by: 

\begin{equation}
\gamma_{p}(s)= \mathcal{P}[sign(BG_{p}^{s})=sign (BG)].
\end{equation}

Apparently from Equation (3), size of the first mini-batch, $BD^{1}$ is considered a bad predictor for the sign of true gradient because of high variance and average convergence of $\gamma(1) = 0.53$. It means for non-iid data, the convergence stays low with any size of mini-batch. However, for iid data, $\gamma$ grows quickly with the increasing batch size which ultimately increases the number of accurate updates. Therefore, signSGD can work on any batch size for training but becomes inefficient in non-iid environments.  

Based on the above \textit{characteristics} and \textit{limitations} of FL environments, we conclude that any communication-efficient FL algorithm needs to attain the following requirements; 1) robust for non-iid environments, 2) scalable for dense networks, vigorous for partial participants and 3) communication compression in both directions; upstream and downstream.
The detailed comparison of FL algorithms based on these requirements is given in Table II. (\textbf{Note:} In Table II, we call "Feasibility for Non-IID Environment" if the FL training converges in non-iid data. We call a "Upstream and Downstream Compression" if the algorithm supports compression in either direction. We call "scalability and partial participation" if the algorithm achieves the desired accuracy in a highly dense distributed network).

\begin{table}
\begin{center}
\begin{tabular}{|p{2.8cm}|p{1.7cm}|p{1.5cm}|p{1.5cm}|}
\hline
   FL Algorithms & Upstream and Downstream Compression & Scalability and Partial Participation  & Feasibility for Non-IID Environment  \\\hline
 \hline
DGC \cite{lin2017deep}, Gradient Dropping \cite{aji2017sparse}, Storm, Variance based \cite{strom2015scalable} & Upstream & Weak & Yes\\ \hline
TrenGrad \cite{wen2017terngrad}, ATOMO \cite{wang2018atomo}, QSGD \cite{alistarh2017qsgd} & Upstream  & Weak & No \\ \hline
SignSGD \cite{bernstein2018signsgd} & Both & Strong & No\\ \hline
Federated Averaging \cite{mcmahan2016communication}& Both & Strong & No\\ \hline

  \end{tabular}
  \end{center}
  
\label{Comparison}
\caption{Comparison of various state-of-the-art algorithms for communication-efficient DL.}
\end{table}

\section{Communication-Efficient Algorithms}

In FL settings, the communication rounds between the server and participant devices are repeated till achieving desired accuracy. However, in the DL models of CNN and MLP training involves millions of parameters that results in high communication-cost. Moreover, the above mentioned limitations cause the delay in uploading the updates by participants. Therefore, the following approaches are proposed to minimise this delay and are considered for evaluation. 

\begin{itemize}

\item \textbf{\textit{FedAvg:}} Federated Averaging (FedAvg) algorithm proposed in \cite{mcmahan2016communication} is evaluated on CIFAR-10 and MNIST datasets. The authors considered two different methods for on-device computation: (i) \textit{parallelism} that enables maximum device participations, (ii) \textit{computation per participant} that enhances local updates to achieve global aggregation. Based on the simulation results, parallelism shows no significant improvement after a certain threshold whereas computation per participant increases the accuracy by keeping a constant fraction of selected participant devices.

\item \textbf{\textit{STC:}} Sparse Ternary Compression (STC) proposed in \cite{sattler2019robust} to meet the general requirement of FL. The authors extend the top-k gradient technique with downstream compression. Considering IID and non-IID environments, results are evaluated compared to the FedAvg algorithm. The simulation results are taken using the CIFAR-10 and MNIST datasets for better comparison and show that STC enables a significant improvement in non-IID environments than FedAvg algorithm. 

\item \textbf{\textit{CMFL:}} Communication-Mitigating Federated Learning (CMFL) proposed in \cite{wangcmfl} to guarantee the global convergence by reducing the communication cost through uploading only relevant local update. To identify the relevancy of each update, participant devices are required to compare the local update with the global update in each iteration. Finally, global update will be done based on the computed relevance score, this global update is considered irrelevant when the relevant score is less than a certain threshold. Simulations experiments conducted on LSTM and MNIST datasets which shows that CMFL requires 13.97 and 3.47 times fewer iterations to achieve 80\% accuracy as compared to FedAvg algorithm. Comparing with Gaia \cite{hsieh2017gaia}, CMFL achieves better accuracy in fewer communication iterations. 

\item \textbf{\textit{FedMMD:}} To alleviate the communication problem in non-IID environments, Federated Maximum and Mean Discrepancy (FedMMD) is proposed in  \cite{yao2019federated}. During training, each participant learns from the other participants to fix the global model by incorporation of MMD into the loss function. Using MMD loss function, the participants are eligible to get more generalised features from global model which accelerates the convergence while reducing communication iterations. The simulation results are taken using CIFAR-10 and MNIST datasets which shows that accuracy is achieved in 20\% fewer communication rounds as compared FedAvg in non-IID environments. 

\item \textbf{\textit{Fed-Dropout:}} To reduce the server-to-participant costs, a lossy compression and federated dropout (Fed-Dropout) approach is proposed in  \cite{caldas2018expanding}. The Fed-Dropout approach uses activation functions for pre-defined number of iterations at fully connected layer to derive a sub-model which is received by the participants for training. Then, the updated sub-model is sent back towards the global model for obtaining a complete DNN model. Fed-Dropout succeeds not only in reducing the communication cost from server-to-participant but also reduces the size of update from participant-to-server. The simulation experiments are conducted on CIFAR-10 and MNIST datasets. Their results show that Fed-Dropout achieves 25\% dropout rate for weight matrices on fully-connected layer and 43\% size reduction in model communication.

\end{itemize}

\section{Algorithm Evaluations}

Based on the aforementioned analysis, none of these algorithms proves the best possible solution. Therefore, they are considered for evaluation in order to find the gaps, issues and limitations which help us to propose a better FL algorithm. 

For evaluations, we considered the following settings which are shared among all the algorithms mentioned in Section III. 

\begin{figure}
  \centering%
    \includegraphics[width=.5\linewidth]{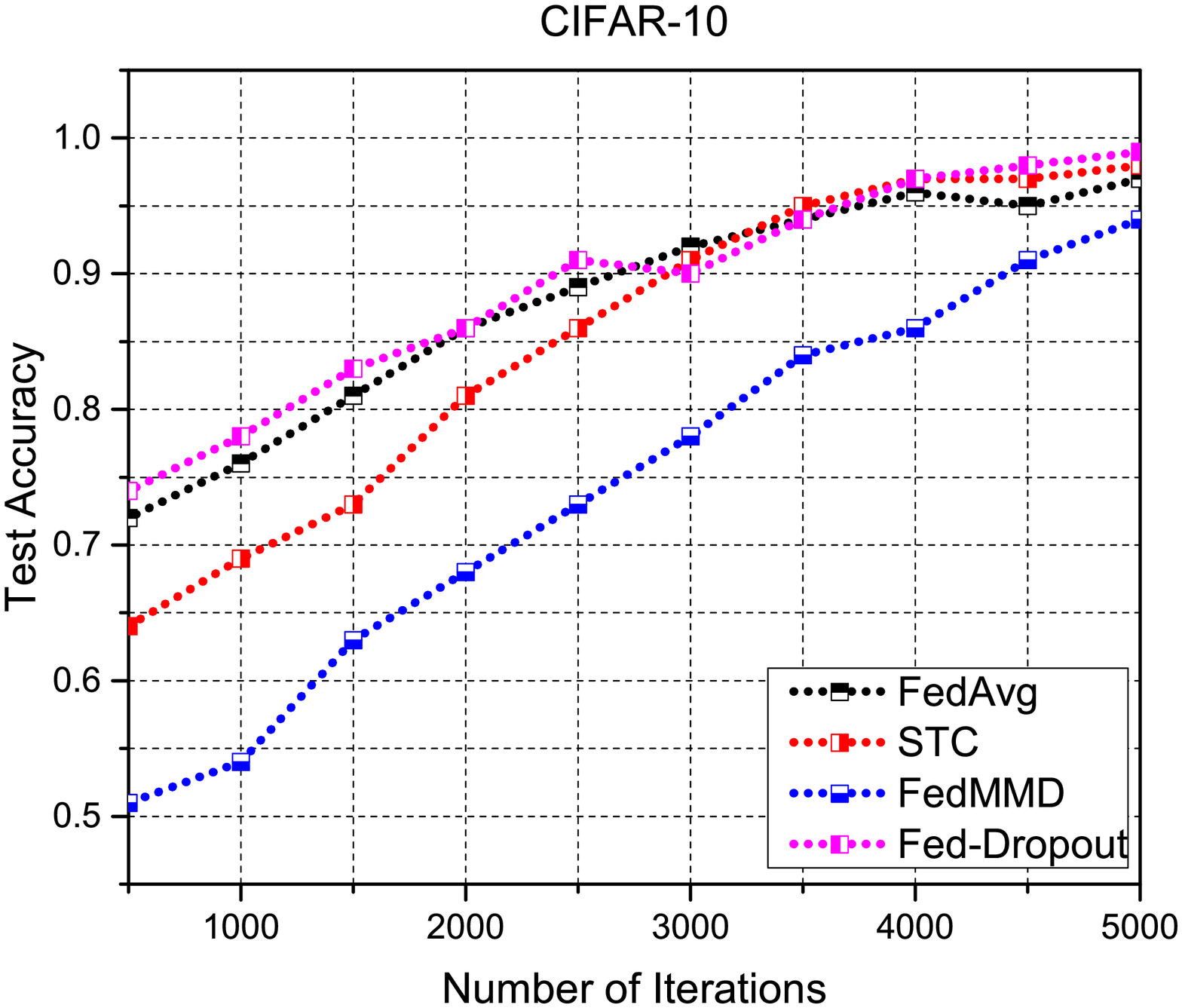}\hfill%
    \includegraphics[width=.5\linewidth]{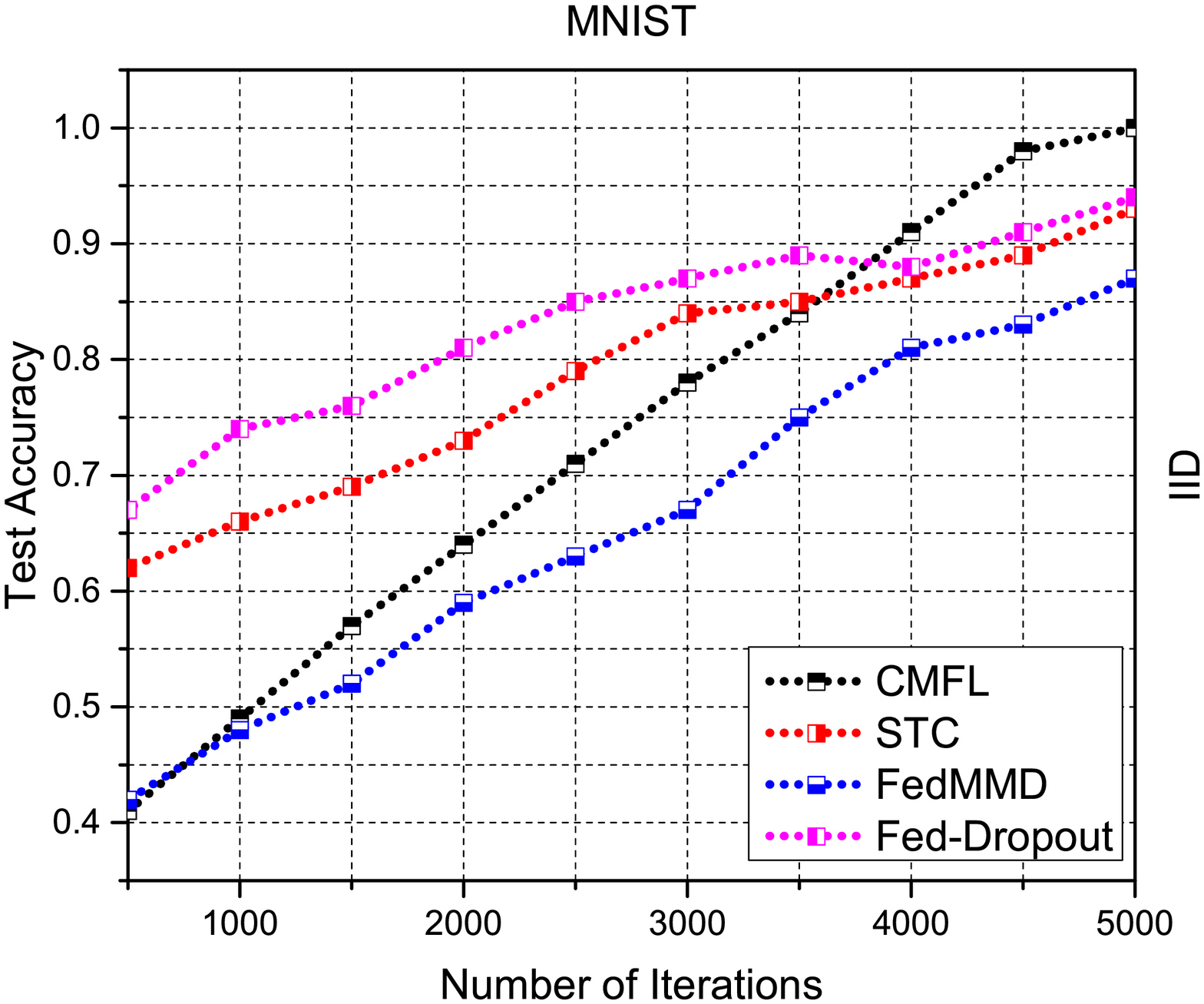}%
    \caption{Test-accuracy comparison w.r.t number of iterations on CIFAR-10 and MNIST datasets for IID environment.}
\end{figure}

\begin{table}[h]
\begin{center}
\begin{tabular}{|p{1.1cm}|p{1cm}|p{1.5cm}|p{1.1cm}|p{1.5cm}|}
\hline
   Parameters & Clients & Participation & Classes & Batch Size  \\\hline
 \hline
Value  & i = 100& P = 10\% & C = 10 & S = 20  \\ \hline

  \end{tabular}
  \end{center}
  \label{Comparison}
\caption{Federated baseline configuration considered in each of the aforementioned algorithms and we utilise these configuration for experimental evaluation.}
\end{table}

\begin{figure}
  \centering%
    \includegraphics[width=.5\linewidth]{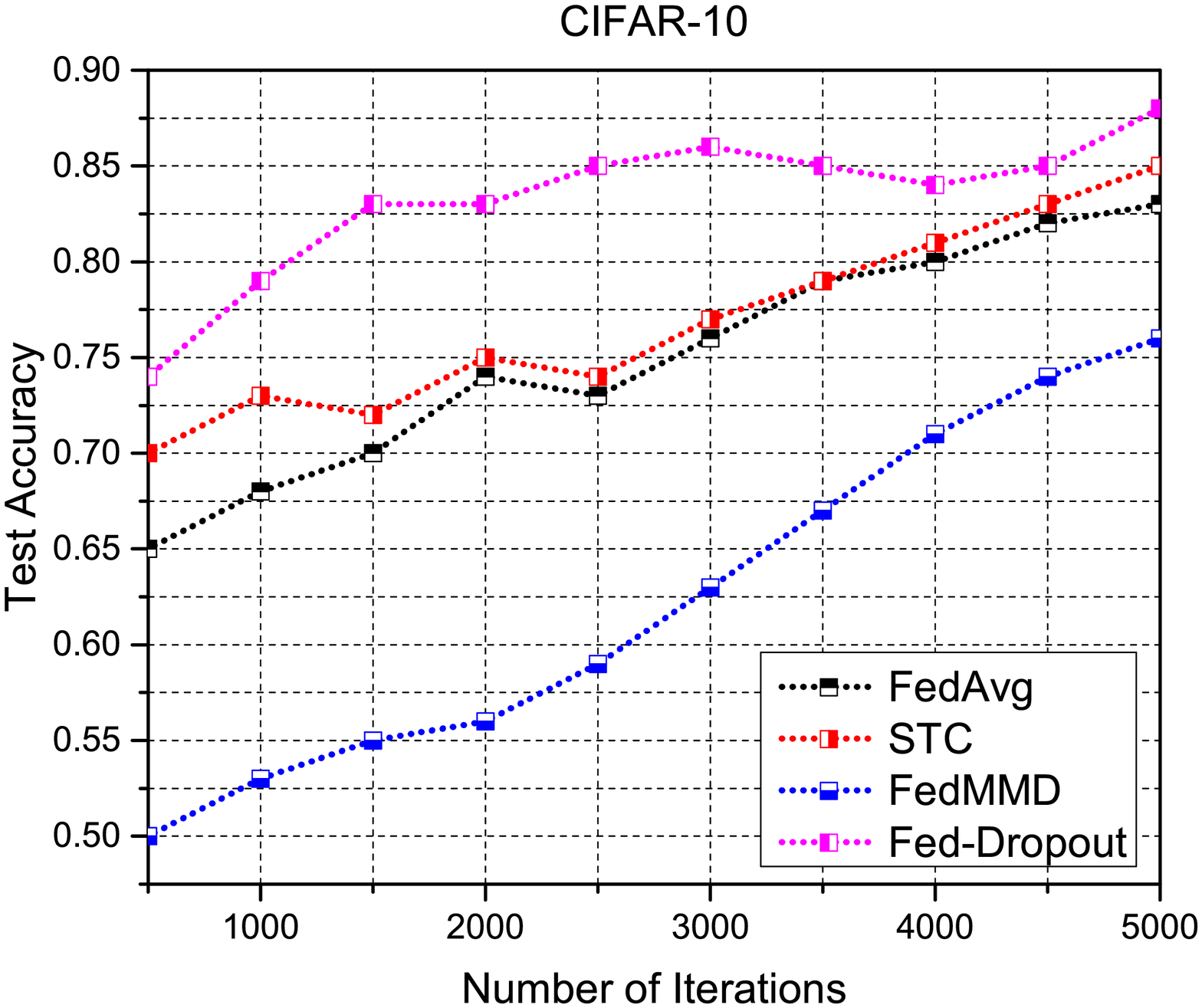}\hfill%
    \includegraphics[width=.5\linewidth]{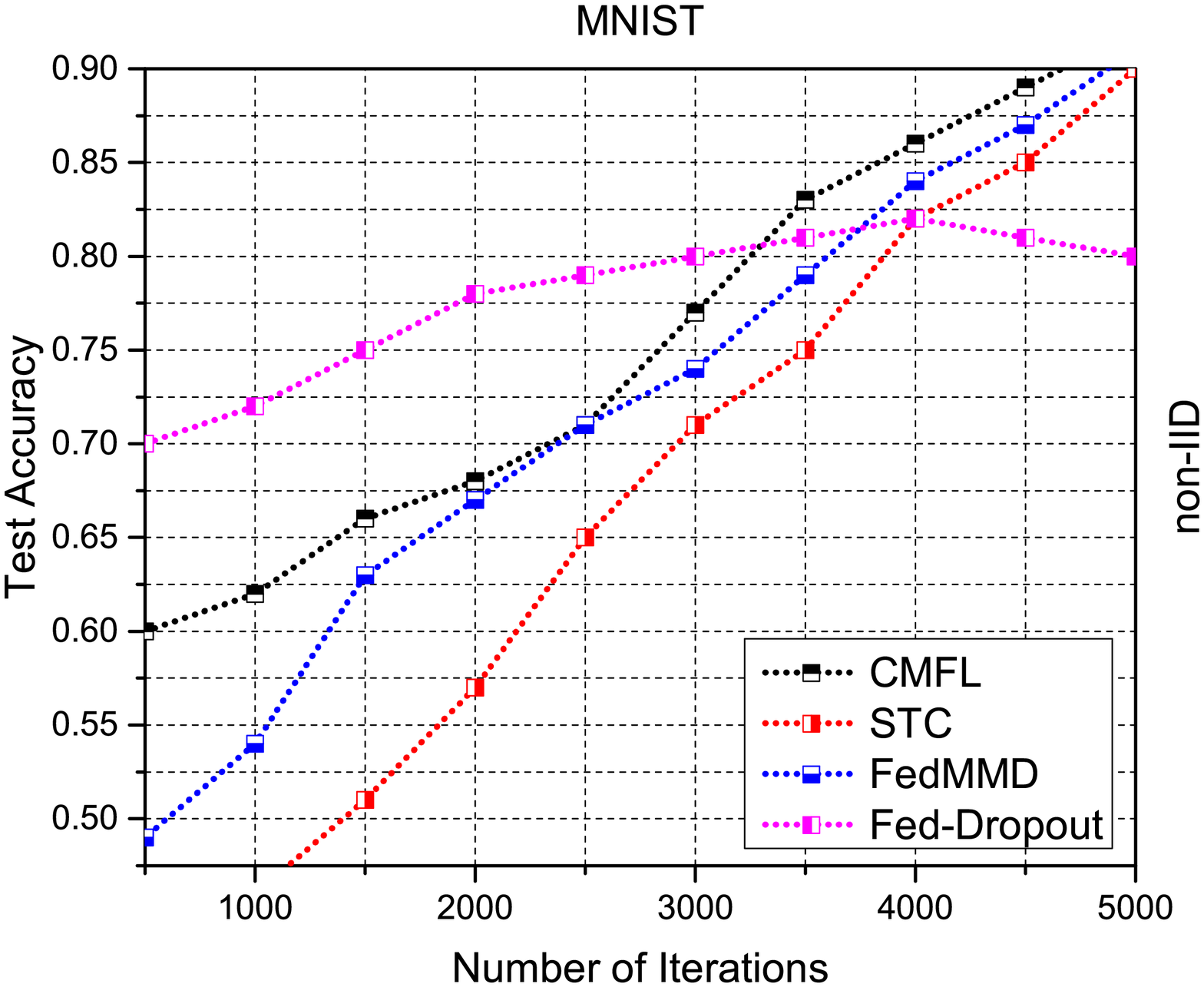}%
    \caption{Test-accuracy comparison w.r.t number of iterations on CIFAR-10 and MNIST datasets for non-IID environment.}
\end{figure}

\begin{table}
\begin{center}
\begin{tabular}{|p{1.5cm}|p{1.9cm}|p{2.1cm}|p{1.7cm}|}
\hline
   Algorithms & Model & Datasets  & Task \\\hline
 \hline
FedAvg  & CNN& CIFAR \& MNIST & Learning Rate,\\ \cline{1-3}
STC  & VGG11 \& CNN& CIFAR \& MNIST & Iterations, \\ \cline{1-3}
CMFL  & CNN& CIFAR \& MNIST & Momentum, \\ \cline{1-3}
FedMMD  & CNN \& MLP & CIFAR \& MNIST& Accuracy \& \\ \cline{1-3}
Fed-Dropout & CNN& CIFAR \& MNIST & Parameters \\ \cline{1-3}
\hline
  \end{tabular}
  \end{center}
  \label{Comparison}
\caption{Hyper-Parameters and Models on Non-IID Data :-learning rate kept constant throughout training process}
\end{table}

The baseline of FL environment is determined by four parameters. In specific, we set the base number of participant clients to 100 with 10\% of participation ratio. Each client is assigned the same number of 10 different classes for training data. In Figures 2 and 3, all hyper-parameters are set to these baseline values as shown in Table III. In addition, the learning tasks and model classification for each algorithm is summarised in Table IV.

On the other hand, we have considered two ways of dividing the CIFAR-10 and MNIST datasets over the clients: 1) \textbf{IID}, where the data is first shuffled and then divided among 100 clients, and 2) \textbf{non-IID}, where the data is first sorted according to their labels and then evenly divided among 100 clients. Figures 2 and 3 show the comparison result of the accuracy of these algorithms \cite{mcmahan2016communication, sattler2019robust, wangcmfl, yao2019federated, caldas2018expanding}  both in IID and non-IID environments, respectively.

In specific, Figure 2, shows the results of test-accuracy in IID environment. Due to the shuffling the data among clients, the accuracy of all four algorithms increases gradually but decay in amount of computation in later stages of convergence. This result shows that Fed-Dropout achieves higher accuracy in CIFAR-10 dataset and comparatively significant accuracy in MNIST dataset. Meanwhile, STC provides the same level of accuracy as Fed-Dropout at later stages of convergence in both CIFAR-10 and MNIST datasets. FedAvg is only considered in CIFAR-10 dataset and achieves better accuracy than FedMMD. Similarly, CMFL is considered only in MNIST dataset and achieves the highest level of accuracy whereas, FedDMMD achieves the lowest level of accuracy in both CIFAR-10 and MNIST datasets. The reason behind this is that in FedMMD each participant is required to learn from other participants which creates higher latency and results in minimum accuracy. 

On the other hand, Figure 3 shows the test-accuracy of the aforementioned algorithms in non-IID environments. In specific, the results show that, in CIFAR-10 dataset Fed-Dropout achieves highest accuracy. Because Fed-Dropout utilises the fully connected layer and obtain a complete DNN model which helps in reducing the cost for both participants and server. Although performing good in CIFAR-10 dataset, Fed-Dropout couldn't get higher accuracy in MNIST dataset. Afterwards, STC performs much better than FedMMD and Fed-Dropout in CIFAR-10 dataset. Meanwhile, FedMMD achieves the lowest accuracy in CIFAR-10 dataset but works far better in MNIST dataset. CMFL achieves the highest accuracy in MNIST dataset by utilising the only relevant update feature.

\section{Proposed Future Work}


Based on the evaluation and analysis presented in this paper, it has been shown that training DL models using FL enhances the security, privacy, and reduces the communication costs. In the future work, we will mainly focus on non-IID environments, not only to reduce the communication costs, but also to provide the enhanced security features. Considering training \textit{MLP} and \textit{CNN} on \textit{CIFAR-10}, \textit{MNIST} and \textit{FEMNIST} datasets, we are hopeful to achieve better accuracy on the aforementioned datasets. Towards this end, we propose a data sharing strategy where the global data model aggregates from uniformly distributed stores. In specific, in the initialisation phase, the global model is trained on the global shared data where a small amount $\Gamma$ of this shared data is distributed among all the participants from $x_1$ to $x_N$. Once all the participants update their shared data portion, then the server aggregates the local model to train the global model. In general, we will consider two tradeoffs; (i) the tradeoff between the size of global model and test accuracy, (ii) the tradeoff between test accuracy and the amount of shared data $\Gamma$. For improving the security, we plan to  combine the secure multiparty computation (SMC) and differential privacy (which are proposed separately in all the existing literature). Since the number of participants is high in FL, we utilise this combined technique to increase the trust-rate among participants. Therefore, we assume that the proposed model will be scalable and able to provide the highest security against threats with the maximum accuracy in communication. 

\section{Conclusion}

In this paper, we present an evaluation of the communication efficiency in FL. We begin with introduction to FL and the need for communication efficient algorithms and explains that FL can play a vital role in promoting the security features while reducing communication cost of mobile devices. Then, we describe the fundamentals of FL by revealing the challenging characteristics of the available FL frameworks. Afterwards, we provide a detailed review of the state-of-the-art FL models, datasets and algorithms. Furthermore, we provide a detailed statistical and experimental evaluation of the existing FL algorithms. This motivates us to propose a novel FL strategy as our future work to improve both security and communication aspects.

\nocite{*}
\bibliographystyle{IEEEtran}
\bibliography{IEEEabrv,IEEEexample}

\end{document}